# A system on chip for melanoma detection using FPGA-based SVM classifier

Shereen Afifi✉, Hamid GholamHosseini
*Electrical and Electronic Engineering Department, Auckland University of Technology Auckland 1010, New Zealand*
{safifi, hgholamh}@aut.ac.nz

Roopak Sinha
*Department of IT and Software Engineering Auckland University of Technology Auckland 1010, New Zealand*
rsinha@aut.ac.nz

**Abstract**

*Support Vector Machine (SVM) is a robust machine learning model that shows high accuracy with different classification problems, and is widely used for various embedded applications. However, implementation of embedded SVM classifiers is challenging, due to the inherent complicated computations required. This motivates implementing the SVM on hardware platforms for achieving high performance computing at low cost and power consumption. Melanoma is the most aggressive form of skin cancer that increases the mortality rate. We aim to develop an optimized embedded SVM classifier dedicated for a low-cost handheld device for early detection of melanoma at the primary healthcare. In this paper, we propose a hardware/software co-design for implementing the SVM classifier onto FPGA to realize melanoma detection on a chip. The implemented SVM on a recent hybrid FPGA (Zynq) platform utilizing the modern UltraFast High-Level Synthesis design methodology achieves efficient melanoma classification on chip. The hardware implementation results demonstrate classification accuracy of 97.9%, and a significant hardware acceleration rate of 21 with only 3% resources utilization and 1.69W for power consumption. These results show that the implemented system on chip meets crucial embedded system constraints of high performance and low resources utilization, power consumption, and cost, while achieving efficient classification with high classification accuracy.*

## 1. INTRODUCTION

Support Vector Machine (SVM) classifier is a common supervised machine learning tool which is widely used for efficient classification. SVM demonstrates high classification accuracy with numerous applications such as speech recognition, object detection, image classification, bioinformatics, medical diagnosis, etc. [1]. Supervised learning machines are typically composed of two main phases, training/learning phase and classification phase.

The SVM training phase constructs a model to be used for classifying any test data that is based on Support Vectors (SVs). The SVs are identified from the training dataset during the training process, to be then used in the classification phase for predicting the proper class of an input test data. SVMs have shown high classification accuracy rates outperforming other popular classification algorithms in numerous cases and applications [2,3].

A growing interest exists for exploiting SVMs in many embedded detection systems and various image processing applications.

The SVM model is computationally expensive and time-consuming especially for large-scale problems, which raises a vital need for acceleration. While software implementations of SVM produce high accuracy rates, they cannot efficiently meet real-time embedded systems constraints. In such embedded real-time applications, special dedicated hardware implementations (accelerators) are required to meet constraints like limited resources utilization and low power consumption. This has motivated plethora of research towards implementing and accelerating SVM in hardware such as using parallel computing platforms.

Special-purpose (reconfigurable) hardware is exploited for boosting computations, while providing High Performance Computing (HPC) at low cost and power consumption. Field-Programmable Gate Array (FPGA) is a robust parallel processing reconfigurable device. FPGA is widely used for realizing essential performance for embedded systems, as well as providing low hardware resource utilization and low power consumption [4]. FPGAs have demonstrated high performance with various applications, which outperformed other comparable platforms [5,6]. Accordingly, FPGA is a suitable platform for realizing an optimized embedded SVM classifier on chip.

Some existing research works aim to implement the SVM model on the FPGA platform. Nevertheless, meeting vital constraints of embedded systems as high performance and low cost are very challenging, in addition to reaching an effective classification system that offers high accuracy rate.

Therefore, this research aims to propose an optimized FPGA-based SVM classifier and implement an embedded classification system on a chip to be used for melanoma detection as a case study. Melanoma is the most aggressive form of skin cancer responsible for the majority of skin cancer-related deaths. The highest rates of melanoma in the world exists in New Zealand and Australia. Early diagnosis of melanoma could reduce mortality rates and treatments costs. Consequently, a real-time embedded classifier is essential for enhancing early detection of melanoma, which could be embedded within a low-cost and fast handheld scanning device for the primary health care.

This study was conducted based on previous experiments performed for melanoma detection within our research group [2]. It was found that the SVM classifier performed better among common classifiers with higher accuracy results for classification and diagnosis of melanoma [2]. In this paper, we propose a hardware/software system on chip for implementing an optimized SVM classifier on FPGA with a use-case on melanoma detection. This article builds significantly on our previous work [7,8]. An initial design and implementation of the system [7], and an early hardware design of the SVM algorithm [8] have been integrated and fully developed into a system on chip in this article. Unlike previous work reported on the implementation of only one model [8], we used the proposed hardware/software co-design to implement three variable-sizes SVM models using different optimization techniques. This article reports extensive results analysis and validation for the three implemented models, as well as comparisons with our previous designs and other related works.

## 2. RELATED WORK ON FPGA-BASED SVM IMPLEMENTATION

Different FPGA-based hardware architectures have been implemented in the literature for realizing the SVM classification phase on FPGA [9]. From reviewing existing implementations in the literature, we concluded that the main challenge is meeting vital embedded system constraints of flexibility, scalability, and high performance, as well as, low cost, area, and power consumption, while achieving effective classification. Many of the current architectures and implementations did not take these constraints into account (especially the critical power constraint that was measured for only a limited number of previous implementations). Most existing implementations were realized on old generations of FPGAs. No FPGA implementation of the SVM exists in the literature that exploits the hybrid architecture (hardware/software system) of the recent FPGA platform "Zynq System on Chip (SoC)" (to the best of our knowledge). Also, almost all previous FPGA implementations are designed utilizing the classical Hardware Description Language (HDL), which is very time consuming and demands expert hardware developers. However, the modern UltraFast High-Level Synthesis (HLS) design methodology is lately exploited for simplifying the FPGA development [10]. Furthermore, no SVM classification system on FPGA exists in the literature that targets early detection of melanoma using clinical images at the primary healthcare.

Consequently, this research focuses on implementing an optimized SVM classifier on FPGA, aiming to overcome such limitations, challenges, and research gaps identified from the performed survey study [9]. A hardware/software co-design is proposed in this paper to implement an embedded SVM classification system for melanoma detection on the hybrid Zynq SoC utilizing the latest HLS design methodology, while meeting the challenging embedded system constraints.

## 3. PROPOSED SVM DESIGN AND IMPLEMENTATION ON SOC

### 3.1. FPGA platform and system development tools
The FPGA platform "Xilinx Zynq-7000 All Programmable System on Chip (SoC)" is utilized to implement our SVM classifier, exploiting the cutting-edge technology and reach a powerful efficient embedded system [11]. The Zynq SoC is characterized by its hybrid architecture, which significantly simplifies the embedded system development process. The FPGA and ARM processor are both combined in a single system on a chip as a Programmable Logic (PL) and a Processing System (PS) respectively.

The software tool "Xilinx Vivado Design Suite" is selected as being an efficient system-design tool for simplifying embedded system development based on incorporating an FPGA within a single SoC [12]. Xilinx Vivado suite comprises a powerful design tool, which employs the new UltraFast HLS design methodology. This methodology is characterized with simplifying FPGA programming via using the High-Level Language (HLL) replacing the traditional HDL [10], as well as, decreasing the FPGA development effort and time.

The following sub-sections present the proposed hardware/software co-design for implementing an SVM classifier on Zynq SoC using the HLS design methodology. The hardware design is first proposed in Section 2 for implementing the SVM classifier as an HLS IP on the Zynq PL part using the Vivado HLS tool, and then the designed IP is integrated into a proposed SoC design in Section 3 using the Vivado design tool. Finally, the software design is proposed in Section 4 using the Xilinx Software Development Kit (SDK) tool to implement the software program running on Zynq PS part and realize the embedded SVM classification system on Zynq SoC.

### 3.2. Proposed SVM HLS IP on Zynq PL
A hardware design is proposed utilizing the recent HLS design methodology to implement an HLS IP of a binary SVM classifier incorporating a linear kernel function. This HLS design/IP implements the SVM classification algorithm, where the main decision function (1) is implemented for classifying a test data sample *x*. Eq. (1) depends on some parameters (α, y and b) that are identified from the training phase, as well as, the number of SVs (denoted as *SV*)[13].

$$F(x) = sign\left(\sum_{i=1}^{SV} \alpha_i y_i (\vec{x}_i \cdot \vec{x}) - b\right) \quad (1)$$

Based on our previous hardware/software co-design implemented in [7], a hardware design extension is

```
Define number of features+1 as F and SVs+1 as SV
FOR each SV
        FOR each feature of the SV
            Read streamed data
            Convert it to float
            Store into array_SVs [SV][F]
        END FOR
END FOR
Read streamed data
Convert it to float
Store into array_αy [0] (b value)
FOR each SV
        Read streamed data
        Convert it to float
        Store into array_αy [SV]
END FOR
FOR each feature
        Read streamed data
        Convert it to float
        Store into array_test [F]
END FOR
FOR each feature
        Clear array_AC [F]
END FOR
FOR each SV
        FOR each feature of the SV
            array_AC [F] += array_αy [SV] * array_SVs [SV][F]
        END FOR
END FOR
FOR each feature
        Distance_value += array_AC [F] * array_test [F]
END FOR
Distance_value -= b
IF (Distance_value ≥ th) THEN
        RETURN 1
ELSE
   RETURN -1
END IF
```

Fig. 1. Proposed pseudo code of the SVM algorithm.

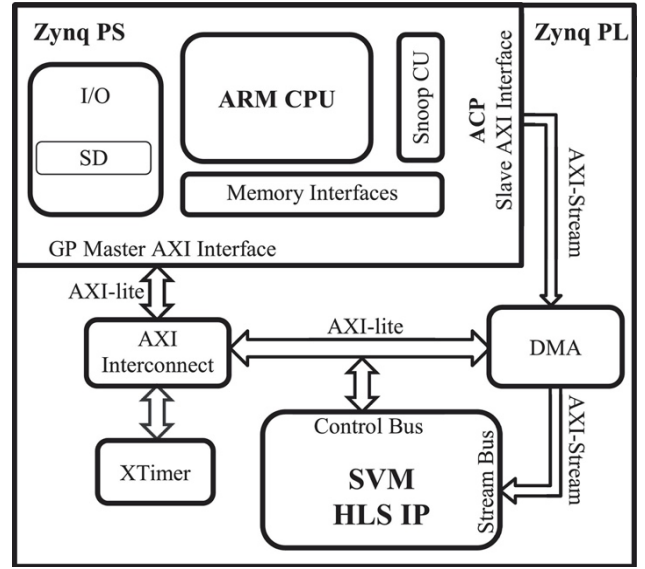

Fig. 2. The proposed hardware/software co-design on Zynq SoC.

proposed in this paper to implement a full SVM onto FPGA. In the initial implementation [7], the most compute-intensive task of the SVM classification algorithm, the dot-product calculation was implemented and boosted on FPGA. So, the hardware design is extended in this paper to reach an online full SVM classifier running on the Zynq platform, offering HPC and low cost.

By exploiting the Vivado HLS tool with C/C++ language, a top function module is designed as an HLS IP that computes the decision function (1), which is divided into three main equations for simplifying the hardware design and mapping, aiming to reduce the hardware complexity.

$$\overrightarrow{AC} = \sum_{i=1}^{SV} \alpha_i y_i \overrightarrow{x_i} \qquad (2)$$

$$D = \overrightarrow{AC} \cdot \overrightarrow{x} \qquad (3)$$

$$F(x) = sign(D-b) = \begin{cases} -1, & (D-b) < th \\ 1, & (D-b) \geq th \end{cases} \qquad (4)$$

The designed IP basically implements the proposed pseudo code that is illustrated in Fig. 1. In the designed C code, float data type is assigned for all used data and mapped to the standard single-precision floating-point format on the FPGA. The designed IP depends on the size of both the features and SVs, in order to implement any SVM classifier. The designed IP receives needed data via an input stream interface to be stored in three main arrays. One 2D array holds features of SVs. The other 1D array has α$y$ of each SV and the third 1D array is for storing the features data of the test instance. The required calculations are divided into three main tasks in the algorithm, mapping the three main Eqs. (2)–(4). The first task is summing all manipulated/multiplied SVs to be stored then in an accumulated array (AC in (2)). The second task is performing the dot product between the test instance and the accumulated array to calculate the classification distance value (D in (3)). Then, the calculated distance number is finally classified according to the sign value, which gives the final SVM classifier decision for the class (F(x) in (4), th is the threshold value determined through the validation phase). We have two possible outputs equal to 1, or −1, which corresponds to melanoma class, or non-melanoma class, respectively.

The HLS tool provides various directives to be applied for the IP to assign different interfaces and apply other hardware techniques [14]. The AXI4-Stream directive is utilized as the input stream interface of the designed function module for streaming the required data between the PS (ARM CPU) and the designed HLS IP in the PL slice of the Zynq SoC. Additionally, the AXI-lite bus is allocated as a control bus of the module for controlling the designed IP and other connected cores in the system as well as controlling the data flow of the system through communicating with the ARM CPU (PS).

In addition, the HLS tool provides various optimization techniques as directives to be employed for the designed IP [14]. In order to optimize loops, the pipelining and unrolling methods are used to enhance the throughput and latency.

The unrolling technique creates multiple independent operations instead of a single collection of operations. Unrolling the loops decreases latency but increases hardware resources utilization as well as power consumption, which could be improved by applying partially unrolling technique. Both pipelined and unrolling designs are applied to the designed SVM HLS IP and investigated based on synthesis results, aiming to find an optimized hardware solution (results are provided and analyzed in Section 4.3).

Concerning arrays, some performance bottlenecks are often added with array accesses. The HLS tool normally maps arrays to dual-port memories to improve throughput. Also, arrays could be partitioned or reshaped by the help of the tool in order to improve memory resource implementation and increase throughput. The available partitioning styles are block, cyclic and complete [14]. These available directives are also applied to the proposed IP for more optimization investigation (discussed in Section 4.3).

### 3.3. Proposed embedded system design on Zynq SoC

The proposed HLS IP of the SVM classifier is successfully co-simulated (RTL simulation) and exported as an RTL implementation (packaged IP), after synthesizing the designed code utilizing the HLS tool. Next, the exported HLS IP is incorporated with the proposed design as shown in Fig. 2. Using the Vivado design suite, the exported HLS IP in the Zynq PL is attached to the PS ARM CPU through an ACP (Accelerator Coherency Port), using a Direct Memory Access (DMA) controller core. The ACP is a 64-bit AXI slave interface on the snoop control unit that allows an asynchronous cache-coherent access point directly from the PL part to the PS part with low latency path [15]. The DMA IP controls transferring the data between the HLS IP and the ARM CPU through the AXI4-Stream bus. Besides, an AXI-Timer is exploited for performance comparisons based on the number of clock cycles needed by the IP/cores.

Finally, the designed Zynq SoC is exported for the SDK tool to be tested, after successfully passing the synthesis, implementation and bitstream generation stages in the Vivado design tool.

### 3.4. Proposed software design on Zynq PS

A test bench or a software program has been implemented to test and verify the implemented SVM classifier on the Zynq SoC. The PS ARM CPU is responsible for executing the test bench besides controlling the attached cores/IPs and the data flow in the system. A software program is designed and implemented in C using the SDK tool. Fig. 3 shows the proposed algorithm of the software test bench/program that runs on the PS ARM processor. The test instance and the parameters of the trained SVM model required for computations are imported, using three main files saved in the Zynq SD card. The first file has the support vectors, and the second file includes $\alpha y$ for each $SV$ with the b value. The third file keeps the test data. All imported data are parsed and stored in three main arrays to be streamed to the Zynq IP for further processing.

The same SVM algorithm as proposed in Fig. 1 is executed on the ARM processor in order to compare its software result with the hardware result resulted from the implemented SVM HLS IP running on hardware. Also for comparing the performance, the XTimer IP is exploited to measure the clock cycles of running the SVM algorithm on software/PS and hardware/PL (including the hardware DMA streaming) and reports the hardware acceleration factor.

This proposed system could be easily adapted to any other trained SVM model with the same size of parameters (number of SVs and features) that targets similar applications or more general classification application.

```
Define number of features+1 as F and SVs+1 as SV
1.  Read SVs data Byte-by-Byte from SVs_File on the SD card
2.  Parse data and store in array_SVs [SV][F]
3.  Repeat steps 1 and 2 for b and αy data, αy_File and array_αy [SV]
4.  Repeat steps 1 and 2 for test data, test_File and array_test [F]
5.  Setup and calibrate the XTimer IP
6.  Run the proposed SVM algorithm/function in Fig. 1 on PS ARM
7.  Measure the total execution cycles by XTimer IP for running SVM
    software on PS and getting the software_result
8.  Initialize the DMA IP and flush the caches
9.  Setup and initialize the hardware SVM HLS IP
10. Run the SVM IP on Zynq PL
11. Transfer array_SVs [SV][F] to the SVM IP via DMA
12. Loop and wait for DMA transfer to be done
13. Repeat steps 11 and 12 for array_αy [SV] and array_test [F]
14. Loop until SVM IP is done
15. Read return value hardware_result from SVM IP
16. Measure the total execution cycles by XTimer IP for running the
    hardware SVM IP and DMA transfers (steps 10-15)
17. Compare and print hardware_result and software_result
18. Compare run time for the hardware and software and print the
    acceleration factor
```

**Fig. 3.** Proposed algorithm of the software program running on Zynq PS.

Specifically, all new data required could be easily loaded via the three designed files stored in the SD card. Accordingly, the proposed SoC is feasible to achieve generality, flexibility, and adaptability.

## 4. EXPERIMENTAL RESULTS

### 4.1. Experiments' setups

A common SVM classifier called "SVM-Light" has been studied as a case study to implement our SVM IP on Zynq SoC. The SVM-light is a robust and simple classifier that is available in C implementation and has been used in various classification problems [16]. The modern UltraFast HLS design methodology is utilized to design and implement a binary SVM HLS IP, using the available SVM-Light classification (C/C++) code.

The training phase was done offline on software by exploiting the available SVM-Light windows application, where the default parameters and the linear kernel function were used to generate the trained SVM models. Based on our previous work within our research group for melanoma detection [2], a dataset was used for training that consists of a total of 356 clinical images, including 168 melanoma and 188 benign images. In order to form a features dataset for the training of the SVM model, some selected pre-processing, segmentation and feature extraction (based on HSV color channels) algorithms were applied to the images dataset (512 × 512 pixels). Finally, a new dataset of 356 instances of 27 features each was extracted from the images dataset, to be used for training the SVM models [2]. In order to achieve a higher accuracy for the trained models, the cross-validation technique was utilized in the training phase. Finally after generating a trained SVM model offline, the model data was extracted to implement the trained SVM model on hardware using this proposed hardware design.

The Xilinx Vivado 2016.1 Design Suite was utilized to design, implement and develop our proposed hardware design on the Zynq-7 ZC702 Evaluation Board. The

Table 1. Synthesis results of applying different directives to the proposed HLS IP.

(A) model 1 (248 SVs)

| Directives | Latency | Throughput | BRAM | DSP | FF | LUT |
|---|---|---|---|---|---|---|
| Basic Interfaces | 82,460 | 82,461 | 17 | 5 | 1265 | 2417 |
| Array Resource: BRAM | 82,460 | 82,461 | 19 | 5 | 1137 | 2376 |
| Array Resource: LUT | 82,460 | 82,461 | 0 | 5 | 1393 | 6015 |
| **Pipeline inner loops** | **14,138** | **14,139** | **19** | **5** | **1251** | **2477** |
| Pipeline most loops | 14,129 | 14,130 | 19 | 10 | 2622 | 3999 |
| Pipeline all loops | 14,129 | 14,130 | 30 | 58 | 5460 | 9516 |
| Unroll inner loops | 9876 | 9877 | 28 | 135 | 13,080 | 28,039 |
| **Unroll most loops** | **8366** | **8367** | **29** | **135** | **13,226** | **49,625** |
| Partial Unroll (factor 2) | 78,959 | 78,960 | 19 | 5 | 1266 | 2645 |
| Array partition (cyclic factor 2) | 13,858 | 13,859 | 22 | 10 | 2105 | 3454 |
| Array partition (cyclic factor 8) | 13,827 | 13,828 | 17 | 10 | 6925 | 10,103 |
| **Array partition (cyclic factor 16)** | **9336** | **9337** | **16** | **20** | **11,113** | **12,835** |
| Array partition (block factor 2) | 42,217 | 42,218 | 22 | 7 | 10,916 | 12,821 |
| Array partition (complete) | 23,512 | 23,513 | 27 | 5 | 23,056 | 10,218 |

(B) model 2 (346 SVs)

| Directives | Latency | Throughput | BRAM | DSP | FF | LUT |
|---|---|---|---|---|---|---|
| Interfaces | 114,898 | 114,899 | 33 | 5 | 1271 | 2429 |
| **Pipeline inner loops** | **19,626** | **19,627** | **35** | **5** | **1258** | **2465** |
| Pipeline most loops | 19,617 | 19,618 | 35 | 10 | 2629 | 4048 |
| Pipeline all loops | 19,617 | 19,618 | 30 | 58 | 5467 | 9550 |
| **Unroll inner loops** | **13,698** | **13,699** | **28** | **135** | **13,919** | **29,975** |
| Unroll most loops | 11,600 | 11,601 | 29 | 135 | 13,793 | 63,328 |
| Array partition (block factor 2) | 59,125 | 59,126 | 38 | 7 | 11,004 | 12,921 |
| Array partition (cyclic factor 2) | 19,248 | 19,249 | 38 | 10 | 2117 | 3503 |
| Array partition (cyclic factor 8) | 19,215 | 19,216 | 40 | 10 | 6490 | 10,027 |
| **Array partition (cyclic factor 16)** | **12,960** | **12,961** | **28** | **20** | **11,123** | **12,938** |
| Array partition (complete) | 32,724 | 32,725 | 27 | 5 | 29,334 | 11,276 |
| Interfaces | 114,898 | 114,899 | 33 | 5 | 1271 | 2429 |
| Pipeline inner loops | 19,626 | 19,627 | 35 | 5 | 1258 | 2465 |
| Pipeline most loops | 19,617 | 19,618 | 35 | 10 | 2629 | 4048 |

(C) model S (61 SVs)

| Directives | Latency | Throughput | BRAM | DSP | FF | LUT |
|---|---|---|---|---|---|---|
| Interfaces | 40,885 | 40,886 | 5 | 5 | 1656 | 2548 |
| **Pipeline inner loops** | **3830** | **3831** | **7** | **5** | **1666** | **2511** |
| Pipeline all loops | 3822 | 3823 | 30 | 105 | 13,854 | 16,195 |
| Unroll inner loops | 3343 | 3344 | 29 | 135 | 20,626 | 26,393 |
| **Unroll most loops** | **2653** | **2654** | **27** | **135** | **19,429** | **45,233** |
| Array partition (cyclic factor 16) | 4483 | 4484 | 27 | 19 | 15,447 | 16,498 |

Vivado HLS tool was used first to develop our SVM HLS IP. Then, the developed SVM IP was exported for integration with the proposed Zynq SoC (Fig. 2) that was designed using the Vivado tool. The designed Zynq system was synthesized, placed and routed and finally the bitstream was generated to be exported for the Xilinx SDK tool to run an online classification application on Zynq.

In the next sub-sections, the implemented SVM models are introduced, then experimental results are presented and analyzed.

### 4.2. Implemented SVM models

Three SVM trained models with different sizes have been developed offline (using the SVM-Light windows application) to be used for the hardware implementation using the proposed design, targeting melanoma detection. Three models were generated from training the available features dataset for melanoma with 356 instances of 27 features each. First, the original full dataset was used to generate a trained SVM model "Model 1" with **346** SVs. Then, data scaling and normalization techniques were applied to the original dataset, which generated another trained model "Model 2" with **248** SVs that achieved higher classification accuracy.

Another third model with smaller scale was implemented in order to be used as a case study for performance validation through running on the Zynq SoC, while the other two models were validated using simulation results only (due to the limited size of the available run-time memory). The small-scale model has **61** SVs generated from using part of the available normalized dataset in the training phase (144 instances).

### 4.3. HLS synthesis results

Some available optimization directives of the Vivado HLS tool were employed and tested (as introduced in Section 4.2), for optimizing the proposed HLS IP design. Different experiments were performed to investigate improving and optimizing the synthesis results by applying various optimization directives of the HLS tool, aiming to achieve an efficient and hardware-friendly design with low hardware complexity. Accordingly, selected HLS synthesis results with the assigned directives were presented in Table 1 (a), (b) and (c) for the three models of different SVs numbers and 27 features each. In Table 1, the first column displays the used directive, whilst the next columns present the HLS synthesis results for applying the corresponding directive (the design latency in clock cycles, throughput in clock cycles, and resource utilization). The first row in the tables demonstrates the synthesis results for the default settings of the HLS tool in addition to applying the interface directives of the I/O ports mapping (AXI4 and AXI-lite as explained in the previous section). The subsequent rows are for applying alternative directives besides the interface. The resource allocation is tested for the used arrays (BRAM and LUT). The pipeline or unroll technique is used for inner/most loops. It is recommended to pipeline the inner loops only in the nested loops, aiming to reach the optimum solution and allowing the HLS tool to make required scheduling quickly [14]. Additionally, the different array partition styles are applied with the loop unrolling directive under the same factor in addition to the pipeline.

Table 2. Hardware implementation results of implemented models on Zynq SoC.

| On-Chip Component | Model 1 Utilization (%) | | | Model 2 Utilization (%) | | | Model S Utilization (%) | | Available |
|---|---|---|---|---|---|---|---|---|---|
| | Design 1 | Design2 | Design 3 | Design 1 | Design2 | Design 3 | Design 1 | Design2 | |
| Slice FF Registers | 2898 (2.7) | 13,830 (13) | 9009 (8.47) | 2898 (2.72) | 13,830 (13) | 9009 (8.47) | 3332 (3.13) | 18,054 (16.97) | 106,400 |
| Slice LUTs | 2579 (4.85) | 12,808 (24.08) | 8111 (15.25) | 2579 (4.85) | 12,808 (24.08) | 8111 (15.25) | 2870 (5.39) | 12,371 (23.25) | 53,200 |
| Memory LUT | 204 (1.17) | 161 (0.93) | 582 (3.34) | 204 (1.17) | 161 (0.93) | 582 (3.34) | 212 (1.22) | 189 (1.09) | 17,400 |
| BRAM | 20 (14.29) | 16.5 (11.79) | 16.5 (11.79) | 20 (14.29) | 16.5 (11.79) | 16.5 (11.79) | 6 (4.29) | 16 (11.43) | 140 |
| DSP48 | 5 (2.27) | 135 (61.36) | 20 (9.09) | 5 (2.27) | 135 (61.36) | 20 (9.09) | 5 (2.27) | 135 (61.36) | 220 |
| Total On-Chip Power | **1.756** | 1.824 | 1.851 | **1.758** | 2.125 | 1.842 | **1.686** | 1.766 | |

By applying different optimization directives to the design with the basic interfaces directives (in the first row), the latency and throughput were significantly improved, whilst extra resources were utilized. This is a justification of the existing trade-off between area and data throughput. The latency of the two large models model 1 and 2 of **82,460** and **114,898** clock cycles was significantly decreased by a factor greater than **9x** and **8x,** respectively. Regarding unrolling loops, lower latency of **9,876** and **13,698** cycles of the two models were demonstrated, however, extra resources were allocated (the power-consuming DSP was increased from **5** to **135** for both models). Also, by applying loop unrolling, the best latency of **8,366** cycles was successfully realized for model 1, with utilizing more resources and using almost all available LUTs of **93**%. Similarly for model 2, the lowest latency of **11,600** cycles was reached, however, it is not applicable for implementation due to the excess utilization of available LUTs of **119**%, which was decreased by unrolling inner loops only. However, by applying the array partitioning (cyclic style on a factor of 16), the best latency of **12,960** cycles was achieved for model 2 with fewer resources (**20** DSPs), while a good latency of **9,336** cycles was achieved for model 1 that was not less than unrolling most loops.

Regarding the pipelining method, the lowest area with only **5** DSPs utilization was realized for both models, whilst latency improvement was achieved (**14,138** and **19,617** cycles). By pipelining more loops, the latency was slightly decreased for both models with some increases in the resources utilization (reaching **10** or **58** DSPs). Accordingly, the pipelined design is considered to be more promising for a low-cost solution that offers low area and power consumption with reduced latency and high throughput.

Similarly for the small model S, optimized results were achieved for latency and resources utilization, especially with applying the pipelining and unrolling techniques.

### 4.4. Hardware implementation results

After developing the designed HLS SVM IP, it was integrated into the proposed Zynq SoC for further implementation using the Vivado tool. Based on the investigation of the HLS synthesis results presented in the previous section, the most effective designs of the HLS IP that showed better synthesis results were selected for the Zynq SoC implementation. For the two large-scale models, the pipelined, unrolled and array partitioned designs (bolded in Table 1 (a) and (b), to be denoted as design 1, 2, and 3, respectively) were implemented (using 100 MHz and 666.67 MHz operating frequency for PL/FPGA and ARM CPU respectively), for being the three best solutions for a balanced trade-off between achieving high-performance and low-area/cost. Also, model S was implemented (using 250MHz frequency for both FPGA and ARM CPU) with the pipelined and unrolled designs (bolded in Table 1 (c)).

#### 4.4.1. Hardware resource utilization

Table 2 summarizes the FPGA resource utilization for the three implemented Zynq systems. The first column shows the hardware resource and then the following columns illustrate the resource utilization's value and percentage value for each of the implemented design. The last column indicates the number of available hardware resources in the target device. It is clear that the percentages of all resources utilization in Table 2 for all design implementations are very low, showing significant improvement in area savings. Regarding the large-scale models, the number of DSPs utilization is equal for the two models, which shows the highest rate of utilization for design 2 compared to other resources (similar behavior for model S).

#### 4.4.2. Power consumption

The power consumption of all implementations have been reported using the Vivado tool (the confidence level is medium) and displayed in the last row of Table 2. The on-chip total power consumption of all implementations of the three models are considered to be small reasonable values, meeting the critical embedded system constraint. For the two large-scale models, the least power consumption results of **1.756** and **1.758 W** were achieved from design 1 that has the least area. Also having almost equal values of power for the two models of different sizes is promising for implementing large-scales SVMs with low power values. The targeted device consumed 9% for the static power and 91% for the dynamic power consumption, where the Zynq PS dissipated 95% of total dynamic power.

**Table 3.** Timing summary of model S.

| (A) FPGA and ARM at 250 MHz | | | | | | |
|---|---|---|---|---|---|---|
| | | FPGA | ARM | Optimized ARM | Speedup1 (ARM/ FPGA) | Speedup2 (Optimized ARM/ FPGA) |
| Design 1 | Clock Cycles | 3693 | 77,367 | 22,398 | **20.95** | **6.06** |
| | Processing Time (µs) | 14.77 | 309.47 | 89.59 | **20.95** | **6.06** |
| Design 2 | Clock Cycles | 3690 | 77,328 | 22,398 | **20.96** | **6.07** |
| | Processing Time (µs) | 14.76 | 309.31 | 89.59 | **20.96** | **6.07** |

| (B) FPGA at 250 MHz and ARM at 666.67 MHz (Design 1) | | | | | |
|---|---|---|---|---|---|
| | FPGA | ARM | Optimized ARM | Speedup1 (ARM/ FPGA) | Speedup2 (Optimized ARM/ FPGA) |
| Clock Cycles | 2815 | 28,968 | 8431 | **10.29** | **2.995** |
| Processing Time (µs) | 11.26 | 43.45 | 12.65 | **3.86** | **1.12** |

**Table 4**
Processing times of model 1 at 100 MHz.

| Processing Time (µs) | FPGA | ARM | Optimized ARM | Speedup1 (ARM/ FPGA) | Speedup2 (Optimized ARM/ FPGA) |
|---|---|---|---|---|---|
| Design 1 | 141.38 | 3093.78 | 905.85 | **21.88** | **6.41** |
| Design 2 | 83.66 | 3093.78 | 905.85 | **36.98** | **10.83** |

**Table 5**
Processing times of model 2 at 100 MHz.

| Processing Time (µs) | FPGA | ARM | Optimized ARM | Speedup1 (ARM/ FPGA) | Speedup2 (Optimized ARM/ FPGA) |
|---|---|---|---|---|---|
| Design 1 | 196.26 | 4483.53 | 1298.04 | **22.84** | **6.61** |
| Design 2 | 136.98 | 4483.53 | 1298.04 | **32.73** | **9.48** |

Similar figures were realized for the other implementations of the two models. For model 1, design 3 showed slightly higher power than design 2. However, design 2 consumed the highest power dissipation of **2.125W** for model 2, as a result of being the design with the highest utilization rates (The power dissipation by the DSPs was increased from less than 1% in design 1 to 5% of the total dynamic power in design 2 for model 2). For the small-scale model, both designs implementations also showed low levels of area and power consumption. Among the other two models and designs implementations, this model achieved the least power consumption of **1.686W** with design 1.

### 4.4.3. Processing speed and time
The AXI-Timer IP core (Fig. 2) was exploited to compare the total computing time between running the designed code (Fig. 1) of the SVM on ARM processor in PS part and on hardware implemented in PL part of the Zynq SoC. Due to size limitation of the embedded DDR3 memory, the application/program could not run completely at the SDK tool, because of the big data size used by both models 1 and 2. Accordingly, the small-scale model S was used for evaluating the processing speed and time by using XTimer measurements from running the application on Zynq Soc, while model 1 and 2 evaluation was based on timing simulation results.

For running the test program on Zynq SoC for model S using the SDK, only 3693 clock cycles was required for running the hardware IP including data streaming via the DMA (for design 1). However, the embedded ARM processor used 77,367 clock cycles as a total run time of a similar software C coded function. Therefore, a significant acceleration factor greater than **20x** was achieved by using the implemented hardware accelerator/IP (similar acceleration values were achieved for the unrolled design 2). The used operating frequency for both the PS Zynq/ARM processor and PL/FPGA was 250 MHz.

Accordingly, the processing times were 14.77 µs and 309.47µs for the IP and ARM, respectively (approximately equal processing time (15.32µs) for the HLS IP was estimated from the C/RTL co-simulation at the HLS tool). Additionally, the ARM processor could run at the most optimization option to reach 89.59µs, whilst the acceleration factor was still significance to be greater than **6x**. Table 3 (a) shows the values of number of clock cycles and processing time for running on the PL, ARM and optimized ARM at 250MHz in addition to the speedup factors. By using the available maximum operating frequency of 250 MHz and 666.67 MHz for the PL and ARM respectively, an acceleration factor of greater than **10x** was achieved regarding the total number of clock cycles, whilst > 3x was achieved regarding the processing time (Table 3 (b)). In addition, the least processing time of 11.26 µs was achieved with the pipelined design of model S.

Tables 4 and 5 summarize the processing times and speedups values for both designs of model 1 and 2, respectively, at 100 MHz. For the largest implemented unrolled model 1, the highest acceleration factors of **36.98x** and **10.83x** were achieved compared to the embedded ARM without and with optimization. That's shows a promising acceleration to achieve a high performance embedded classification system, while increasing scalability.

### 4.4.4. Classification accuracy
Using the Xilinx SDK tool, the proposed test bench presented in Section 3.4 (Fig. 3) has been developed in C. Some test instances of extracted features were tested to be correctly classified by all implemented SVM IPs (one at a time). Model S has been validated for online classification while running on the Zynq SoC, whilst other implemented models have been verified based on simulation results (due to memory size limitation at run-time).

Table 6. Parameters of implemented SVM Models.

| SVM Model | Training Dataset | | SVs | Accuracy % |
|---|---|---|---|---|
| | Melanoma | Benign | | |
| Model 1 | 168 | 188 | 248 | **80.85** |
| Model S | 100 | 44 | 61 | **97.92** |

Model 1 (248 SVs) was trained using the cross-validation method to produce a model with a good classification accuracy of **80.85%**. In order to verify the hardware classification result of our implemented SVM IP and compare it with the software result, we monitored the calculated classification distance value $D$ in (3). By using the C/RTL co-simulation results from the HLS tool, the distance value from our implemented IP was easily compared to that value generated from the SVM-Light window application/software to be identical for all tested instances. Accordingly, the percentage error was equal to zero, reserving the classification accuracy level without any loss from the hardware implementation, in contrast to some existing implementations in the literature as stated in Section 3.

Same performance was achieved from the small-scale model S with a great accuracy of **97.92%**, which has been validated by both simulation results and online classification results while running on the Zynq SoC. Table 6 summarizes different parameters of the implemented SVM models; model 1 and model S. For each model, the table shows the number of SVs generated with the number of instances used in the training dataset of 27 features each, as well as, the classification accuracy rate that was preserved without any loss and validated with our experiments.

## 5. DISCUSSION AND COMPARISONS

### 5.1. Discussion

The main objective of this research is proposing an optimized FPGA-based SVM classifier towards realizing an efficient embedded classification system targeting early detection of melanoma (as a case study). This research aimed to contribute to the existing literature by considering the existing challenges, limitations and research gaps in the current literature (that were identified from the survey study [9] and summarized in Section 2). This research focused on meeting the challenging constraints of embedded systems development, while achieving efficient classification with high accuracy rate. Recent FPGA design methods, technologies, and system development tools were exploited for implementing the proposed designs.

By using the modern UltraFast HLS design methodology and the available optimization techniques (directives), the development effort and time were declined, whilst the embedded system design process was simplified. A simple and scalable HLS-based design was proposed for implementing an efficient SVM classifier IP/accelerator on Zynq SoC. The main decision function was divided into three main equations for simplifying the hardware mapping, aiming to reach a hardware-friendly design with lower hardware complexity.

Three trained SVM models with different sizes were implemented using the proposed design, which were generated from the training with the available feature dataset for melanoma detection. In addition, another dataset of different application 'pattern recognition' has been used and tested, in order to validate our previous initial design [7] and this proposed extended design for general purpose classification. The trained SVM model has 877 SVs and 9947 features, which was fully implemented and tested with our previous hardware/software co-design [7]. However for implementing such a large model using this proposed extended hardware design, an FPGA of bigger capacity with more resources is required or an additional device(s) could be co-operated to be fully implemented on hardware. Accordingly, the proposed design is capable of implementing SVM classifiers of various applications with variable sizes. Therefore, generality, scalability, and applicability of classification could be realized with the proposed design.

By applying various optimization techniques of the HLS tool, different hardware solutions were achieved based on HLS synthesis results, aiming to find an optimum solution/balance for the existing trade-off between performance and area/cost. Different designs have been implemented, offering a low-cost design, as well as, a fast design with higher cost. These various design options can be chosen by developers to meet their project requirements. The low-cost design with lower area and power was favored for our case study "melanoma detection", for realizing a low-cost handheld device. For the Zynq SoC implementation, the most effective designs of the HLS IP were selected for implementation that showed the best synthesis results, aiming to reach a balanced trade-off between achieving high-performance and low-area/cost.

The experimental results of the three implemented SVM models on Zynq SoC using the proposed design have been evaluated in this discussion based on these factors; hardware resource utilization, power consumption, processing speed and time and classification accuracy, aiming to reach an optimized hardware solution for melanoma detection, while meeting vital embedded systems constraints.

Regarding the resource utilization, all design implementations of the three different models showed very low utilization percentage that significantly improved area costs. The unrolled design 2 is considered to be the most costly design compared to the other two designs that has the largest area results for almost all resources. The pipelined design 1 demonstrated the least area results for the three models. Therefore, the pipelined design is considered to be the most cost effective design for achieving an embedded system for online classification with low area and low cost.

All implemented designs demonstrated very low power consumption, while the least values were for the pipelined design. So, the pipelined design is capable of meeting the most challenging constraint of "low-power consumption". In addition, it has been observed (from different implementation experiments) that when using less operating frequency for the ARM processor in the Zynq PL part, the power dissipation is decreased. This method could be applied in the future to optimize the power consumption of the other models. Therefore, our implementation of the proposed low-power system is so promising for the deployment in an embedded environment, aiming to reach our ultimate goal of realizing a handheld device with high performance and low cost.

A significant hardware acceleration factor from **20x** to **36x** was achieved by the implemented hardware systems of variable sizes compared to an equivalent software implementation of the SVM classification function running on the embedded ARM/PS processor. The least processing time of 11.26μs was achieved with the pipelined design of the small-scale model S at 250MHz. Consequently, a real-time embedded SVM system can be achieved that is scalable and easily extended offering high performance.

Regarding the classification accuracy, the experimental results demonstrated that every hardware classification result of the implemented classifiers was exactly equal to the corresponding software classification result. Accordingly, the accuracy rate was preserved without any loss from our hardware proposed design, in contrast to some existing implementations in the literature that suffered from slightly loss in the accuracy rate. A great accuracy of **97.92%** was achieved by model S, which has been validated by both simulation results and online classification results while running on the Zynq SoC. Therefore, a reliable scalable online SVM classification with a high classification accuracy could be realized with no loss in accuracy rate using our proposed design, while meeting critical embedded system constraints of optimized speed, area, power and cost.

Finally, the implemented SVM classifier "model 1" on Zynq SoC with the pipelined design is considered to be an optimized classifier for our application "melanoma detection", which is an efficient trained model of realistic size with only **2.7%** slices utilized and **1.7 W** power consumed, while classifying was achieved at **56 μs** with **80.8%** accuracy at 250MHz.

**5.2. Comparison with our previous proposed designs**

*5.2.1. Comparison with our previous hardware/software Co-design*
The previous hardware/software co-design [7] was proposed in order to implement the complicated dot-product calculation onto the hardware/PL, while the rest of the required calculations of the decision function (1) was running on the software/PS in a single device (Zynq SoC). In this proposal, the hardware design was extended for implementing the whole decision function in the PL as a Zynq coprocessor/accelerator IP. The extended hardware design depends on the size of both features and SVs, whilst the previous design depends on the number of features only.

The implementation of the previous design was realized for the SVM classifier that has the same size as model 2 (27 features and 346 SVs). So, design 1 (pipelined) and design 2 (unrolled) of model 2 were chosen in order to compare the implementation results of the previous design with this extended hardware design. Table 7 shows the implementation results of model 2 implemented on Zynq SoC using our previous hardware/software co-design with the unrolled HLS directive [7] and this proposed extended design (with the pipelined (design 1) and unrolled (design 2) directives). The extended hardware implementation of design 1 successfully extended our previous implementation, while consuming extra **0.02 W** only for power dissipation. For resource utilization, additional 17 BRAMs and 31 memory LUTs were used, whilst utilization of other hardware resources (FFs and LUTs) was reduced with equal number of DSPs. Compared to design 2, an additional **0.387 W** was required than that of the previous implementation, and more resources were utilized for all resources except for Memory LUTs.

Obviously, design 1 showed more optimization in area and power results compared to design 2 for extending the previous design, where the unrolling technique was used. Accordingly, more design extension and scalability could be easily achieved with good implementation results by using the proposed cost-effective pipelined design.

*5.2.2. Comparison with our proposed BRAM-based design*
We have proposed another similar design on [17], which is based on using three BRAM interfaces to pass the corresponding three arrays' data instead of streaming required data via the stream interface/bus. Same SVM trained models, model 1 and model S have been implemented in [17] using the BRAM-based design with applying the pipelined technique/directive to boost required processing. So, the pipelined design 1 of both model 1 and model S using this proposal were compared to the implemented models using our proposed BRAM-based Design [17], where the hardware implementation results are summarized in Table 8.

This proposed hardware/software design consumed less hardware resource utilization with same number of DSPs and lower power consumption for both models, while keeping the same level of the classification accuracy. This shows that our different proposed designs are promising for preserving accuracy without loss, while improving hardware implementation results for reaching an optimum solution. For the small-scale model S, this proposed design consumed 0.4 W less power consumption than the BRAM-based design. Also, fewer hardware resources were utilized, especially for the BRAM utilization where only 6 BRAMs were utilized, while the other design used 48 BRAMs. Similar figures of less resources and power were recorded for the large-scale model 1.

Regarding the processing time, this design spent less 0.2 μs than the other design for the model S at 250MHz, while extra 17.3μs was required for model 1. It can be considered as another justification of the existing trade-off between performance and area/cost. This proposed hardware/software co-design still shows low processing time with fewer resources and less power than the other design for variable model's size. Therefore, this design is feasible to realize an efficient embedded classifier for integration within a cost- and energy-efficient handheld device, targeting melanoma detection at the primary healthcare.

Table 7. Implementation results comparison of model 2.

| Results | | Proposed Design | | Our Previous Design [7] |
|---|---|---|---|---|
| | | Design 1 | Design 2 | |
| Resources Utilization | Slices | 2898 | 13,830 | 5584 |
| | LUTs | 2579 | 12,808 | 4373 |
| | Memory LUT | 204 | 161 | 173 |
| | BRAM | 20 | 16.5 | 3 |
| | DSP48 | 5 | 135 | 5 |
| Power (W) | | 1.758 | 2.125 | 1.738 |

Table 8. Implementation results comparison of model 1 and model S.

| Results | | Model 1 | | Model S | |
|---|---|---|---|---|---|
| | | Proposed Design | BRAM-Design [17] | Proposed Design | BRAM- Design [17] |
| Resources Utilization | Slices | 2891 | 30,006 | 3332 | 10,874 |
| | LUTs | 2566 | 17,506 | 2870 | 7218 |
| | Memory LUT | 204 | 2873 | 212 | 874 |
| | BRAM | 12 | 48 | 6 | 48 |
| | DSP48 | 5 | 5 | 5 | 5 |
| Power (W) | | 1.756 | 2.65 | 1.686 | 2.06 |
| Processing Time (μs) | | 56.6 | 39.3 | 11.26 | 11.46 |
| Classification Accuracy % | | 80.85 | 80.85 | 97.92 | 97.92 |

**5.3. Comparison with related works**

In order to compare with some relevant work for different applications, our proposed pipelined design 1 of model 1 is selected, as being the most effective design with real data size for our case study "melanoma detection". The selected model showed the best implementation results of high performance and low cost, area, and power among the implemented designs and models (apart from the small-scale model S).

*5.3.1. Detection accuracy*

Some FPGA-based implementations in literature suffered from some loss in the SVM classification accuracy as stated in Section 3. Interestingly, no loss in the classification accuracy rate was achieved from our hardware implemented model, as the calculated classification values were exactly equal to the corresponding software results, which ensures reservation of the online classification accuracy rate onto FPGA. That's increase feasibility of implementing other SVM models using our proposed design, while keeping same classification accuracy level without loss. In addition, our implemented SVM classification system showed a high acceptable detection accuracy of more than 80%, which could be used and applied in real life (>97% for model S). Precisely, our implemented system is considered to be accurate and reliable with zero loss in classification accuracy rate in contrast to other reported implementations in existing literature [18–22] (have recorded some loss in accuracy).

*5.3.2. FPGA technology*

The fact that many implementations in literature used old versions of FPGAs and very limited used recent ones [23,24], motivates us to use the latest FPGA technologies that are more powerful and feature-rich. Accordingly, we used the recent Xilinx Zynq-7000 SoC platform for our implementation. It also allowed for a higher operating frequency in contrast to numerous previous implementations in the literature. Besides, the modern Vivado Design Suite (2016.1 version) was exploited for our development process that applies the latest UltraFast HLS design methodology, which decreases hardware development effort, accelerates design productivity and shorten time-to-market. Consequently, our implemented system on Zynq achieved more optimized implementation results.

*5.3.3. Processing speed and time*

Regarding the processing speed and time, the implemented system on hardware has significantly accelerated the processing power up to **36** orders of magnitude over similar software implementation running on the embedded ARM processor/CPU. By using the recent FPGA technology that offers a high operating frequency of 250 MHz, a processing time of 56.6 μs (33.5 μs in case of the unrolled design) was achieved, demonstrating real-time performance (the least time of 11.26μs was achieved for the pipelined model S). Therefore, a real-time and effective embedded system can be achieved, which also outperforms some existing implementations regarding processing time [18,19,25–29].

*5.3.4. Hardware resources utilization*

Regarding the hardware resource utilization, our proposed implementation reported very low utilization of all resources, which ensures the realization of a low power system with low cost and feasibility for extensibility and scalability. Our implemented system demonstrated less resource utilization than some of the previous implementations for different applications in literature [18,20,22,25,26,28,30].

*5.3.5. Power consumption*

Interestingly, our achieved **low-power** embedded system meets the most challenging constraint "low power consumption", whereas very few of such implemented systems exist in the literature. Also, it has been found that most existing implementations had not included any measurements for the power consumption. Specifically, our

Table 9. Comparison with related works.

| Related Works | | Model 1 | Model S | [25] | [30] | [26] | [28] |
|---|---|---|---|---|---|---|---|
| FPGA Recourses | Slice FF Registers | 2891 | 3332 | 59,208 | 23,220 | 5162 | 12,674 |
| | Slice LUTs | 2566 | 2870 | 122,637 | 57,296 | 8887 | 41,135 |
| | BRAM | 12 | 6 | 2049 | 83 | 74 | 132 |
| | DSP48 | 5 | 5 | N/A | 40 | 64 | 64 |
| Power (W) | | 1.756 | 1.686 | 15 | N/A | N/A | N/A |
| FPGA | | Zynq-7000 | Zynq-7000 | Virtex-5 LX220 | Virtex 5-LX110T | Virtex 5-LX110T | Virtex-5 LX110T |
| Features size | | 27 | 27 | N/A | N/A | 400 | 512 |
| Number of SVs | | 248 | 61 | 16 | 74–467 | 818 | 512 |
| Application | | Melanoma Detection | Melanoma Detection | Skin Classification | Object detection | Object detection | Pedestrian detection |
| Kernel | | Linear | Linear | Gaussian | Polynomial RBF | Polynomial | Linear Polynomial RBF |
| Detection Accuracy | | 80.85% | 97.92% | N/A | 76–78% | 88% | N/A |
| processing time | | 141, 56.6 μs (83, 33.5 unrolled) | 11.26 μs (14.76 unrolled) | 0.02 s | N/A | 54 μs | 44.84, 82.5 μs |
| Frequency(MHz) | | 100, 250 | 250 | 200 | 100 | 100 | 50, 92 |
| Hardware Architecture | | HLS-based pipelined (unrolled) | HLS-based pipelined (unrolled) | Fully pipelined | Systolic array | Systolic array | Pipelined |

Zynq systems dissipated lower power among other existing implementations in the literature [20,22,25,31].

*5.3.6. Detailed comparison and discussion*

Some of the existing related implementations of binary SVMs were selected to be compared with our implemented model 1 and model S using the pipelined design 1 (apart from the different applications used), which is summarized in Table 9. Using the modern UltraFast HLS design methodology, our implementations achieved significant hardware results compared to others that used the traditional pipelined architectures and common systolic array architectures. It is clear that lower resource utilization was demonstrated with real moderate size of SVM parameters with the linear kernel. The power consumption is significantly low (less than 1.8 W), compared to a very high power of 15 W in [25], while others didn't consider this critical constraint. By using the recent FPGA platform, a higher operating frequency was used to achieve extremely less processing time than [25] and comparable time to [26] and [28]. Regarding model 1, the least processing time of 33 μs was demonstrated by using the faster unrolled design that offers slightly higher cost and area from the pipelined design, however, the lowest 11.26 μs was achieved by the small-scale model S using the cost- and energy-efficient pipelined design. Acceptable classification accuracy rate higher than 80% with zero loss was verified for our application, while others didn't validate their hardware classifiers [25,28]. Accordingly, our implemented models on the recent hybrid Zynq SoC platform achieved optimized results for the hardware resource utilization, power consumption, detection speed and processing time with high classification accuracy rates using real data for melanoma detection.

Finally, to the best of our knowledge, our Zynq implemented embedded system using the HLS method is considered to be the first FPGA-based SVM classifier exists in the literature that targets melanoma classification. In addition, our implemented system successfully overcame the most challenges exit in the literature of meeting critical embedded system constraints of high performance, flexibility, scalability, and low area, cost, and power consumption, while reaching effective classification.

# 6. CONCLUSIONS

The main contribution of this research is proposing a hardware/software co-design to implement a full SVM classifier on FPGA, and realization of an embedded SoC dedicated for melanoma detection on the latest hybrid Zynq SoC utilizing the modern UltraFast HLS design methodology. Our implemented Zynq systems met the challenging embedded system constraints, by achieving high performance computing at low cost, area and power consumption, while realizing high classification accuracy. By utilizing the HLS design methodology and the offered optimization techniques (directives), the development effort and time were declined, whilst embedded system design process was simplified. A simple flexible IP-based design was presented, which is scalable and easily extendable to support multi-purpose classification. Different solutions/designs were presented for balancing the existing trade-off between speed and area, offering options for various project requirements from a low-cost design to a fast design with higher cost.

Interestingly, the SVM classification process was significantly accelerated on FPGA by a factor up to **36x** outperforming an embedded processor, whereas **11.3μs** processing time was achieved using a high operating frequency of **250**MHz. Moreover, an effective SVM classification with high accuracy of **97.9%** was realized on hardware without any loss in accuracy rate, in contrast to other existing implementations in the literature. Compared to other related systems in the literature, our implemented embedded system is a cost-effective system with low area *(***2.7%** slices) and power consumption (**1.7W**).

For future work, the implemented embedded system of the SVM could be easily extended and adapted for different online classification applications, targeting generality, scalability and applicability. Also, the implemented binary classifier could be easily extended to implement a multiclass classifier, in addition to implementing different types of kernels. More test instances would be tested in future in order to validate the classification accuracy rate of the implemented classifier. Furthermore, the proposed hardware/software co-design realized on the hybrid Zynq SoC using the HLS method could be adopted by hardware developers for implementing their embedded systems. Other FPGA-based design and optimization methods can be employed in the future (e.g. fixed-point arithmetic, DPR technique, multiplier-less method), for gaining more flexibility and scalability with higher performance and less cost. The presented scalable IP-based design would be extended to form a multi-core architecture by adding more SVM IPs in a single device/SoC that could be applied as a multi-class, ensemble, or cascaded classification. Finally, the implemented classifier is feasible to be embedded in the future within a fast low-cost handheld medical scanning device for melanoma detection or any other applications.

## CONFLICT OF INTEREST

The authors declare that they have no conflict of interest.

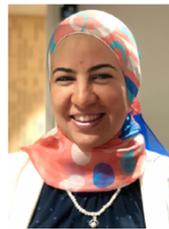

**Dr. Shereen Afifi** was awarded a B.Sc. in Computers and Systems Engineering with honours from Ain-Shams University, Egypt in 2003 and a M.Sc. in Computer Engineering from Arab Academy for Science, Technology and Maritime Transport, College of Engineering and Technology, Egypt in 2012 with excellent grade. She finished her PhD studies in Electrical and Electronic Engineering at the School of Engineering, Computer and Mathematical Sciences, Auckland University of Technology (AUT), New Zealand in 2018, having received an AUT Vice Chancellor's Doctoral Scholarship. The PhD research project aims to propose an optimized hardware-based embedded system to enable a low-cost handheld medical device dedicated for early detection of melanoma, which will be very beneficial for the primary healthcare in New Zealand. Her research interests include FPGAs, Embedded Systems, Microprocessors and Microsystems, Biomedical Devices, e-Health, Machine Learning and Computer Vision. She has been working in the academic career since her graduation as a Lecturer/Teaching Assistant in the French University in Egypt. She has been also a researcher in the Informatics Research Centre-French University in Egypt and successfully completed a research internship in the Laboratory LEAT at The University of Nice Sophia-Antipolis, France in July 2013 having received a scholarship from the French Institute in Egypt. She is currently working in Postdoctoral Research and Teaching in the School of Engineering, Computer and Mathematical Sciences at AUT University.

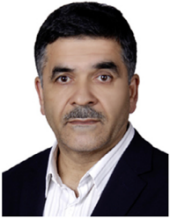

**Dr. Hamid GholamHosseini** is an Associate Professor at the Department of Electrical and Electronic Engineering, Auckland University of Technology, New Zealand. He completed a PhD in Biomedical Engineering at the Flinders University of Australia in 2002 and fulfilled his MSc in Electrical Engineering at the University of Tehran, Iran. His current research and development work combines areas of expertise ranging from Smart patient monitoring/ Biomedical signal and image processing/ Embedded and reconfigurable systems/Odor reproduction and its application to tele-olfaction olfactory systems/ and e-Health for smarter healthcare. Dr. GholamHosseini holds honorary and visiting Professorships at Mälardalen University, Sweden and Guangdong University of Technology, China and has published more than 140 book chapters, journal papers and conference publications. He is a senior member of IEEE and Chapter Chair of Engineering in Medicine and Biology Society (EMBS) of IEEE New Zealand North Section.

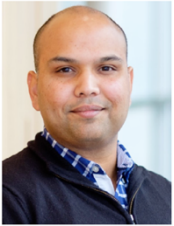

**Dr. Roopak Sinha** received his Ph.D. in electrical and electronic engineering from the University of Auckland, New Zealand. He also holds BE(Hons) and MCE degrees gained in 2003 and 2016, respectively. He has previously worked with Institut National de Recherche en Informatique et Automatique (INRIA), Grenoble, France and The University of Auckland. Currently, he is a Senior Lecturer with the School of Engineering, Computer and Mathematical Sciences, Auckland University of Technology, New Zealand. His research interests include next-generation formal frameworks for designing large-scale embedded software with application in industrial automation systems, Internet-of-Things, and intelligent transportation systems.